\pdfoutput=1

\documentclass{article}

\usepackage{arxiv}

\usepackage[utf8]{inputenc}
\usepackage[T1]{fontenc}
\usepackage{hyperref}
\usepackage{url}
\usepackage{booktabs}
\usepackage{amsfonts}
\usepackage{amsmath,amssymb}
\usepackage{nicefrac}
\usepackage{microtype}
\usepackage{graphicx}
\usepackage{xcolor}
\usepackage{multirow}

\title{When More Modalities Hurt: Modality Dropout for Heavy-Duty Vehicle Engine Diagnostics}

\author{
Adeel Zafar, \quad
S\l{}awomir Nowaczyk, \quad
Hamid Sarmadi, \quad
Saeed Gholami Shahbandi \\[0.5em]
Center for Applied Intelligent Systems Research, 
Halmstad University, Sweden \\
\texttt{\{adeel.zafar, slawomir.nowaczyk, 
hamid.sarmadi, saeed.gholami.shahbandi\}@hh.se}
}

\begin{document}
\maketitle

\begin{abstract}
Heavy-duty vehicle diagnostics generate three 
disconnected data modalities: unstructured 
multilingual service complaints, high-dimensional 
sensor telemetry with over 80\% missing values, 
and Diagnostic Trouble Codes (DTCs). We investigate 
whether fusing these modalities improves engine 
component classification on a proprietary dataset 
from a major truck manufacturer. Through 5-fold 
cross-validation across multiple model 
configurations spanning three model families on 
five engine component classes (885 samples, the 
full cross-database matched population for this 
manufacturer), we find that naive fusion provides 
modest gains over text alone (65.3\%). However, 
modality dropout during training, which randomly 
disables entire modalities per batch, forces the 
network to exploit weaker inputs and achieves 
68.8\% accuracy on text+DTC fusion (weighted F1: 
0.67), a 3.5-point improvement over text-only 
(65.3\%, weighted F1: 0.64) and the best result 
across all methods including logistic regression 
and gradient-boosted trees. Per-class analysis 
shows that the dominant modality varies by fault 
type: text describes symptoms, DTCs encode 
structured fault signals, and sensors measure 
physical state. On intake/exhaust faults, sensors 
alone reach 93\% where text achieves 80\%. 
On fuel system faults, fusion with modality dropout nearly triples accuracy from 15\% to 38\% over text alone. To our 
knowledge, this is the first application of 
three-way modality fusion combining text, sensors, 
and fault codes in industrial vehicle diagnostics.
\end{abstract}


\section{Introduction}

Heavy-duty trucks generate three parallel streams 
of diagnostic data. Technicians write service 
complaints describing observed symptoms. Electronic 
Control Units (ECUs) produce sensor telemetry that 
captures the physical state of vehicle components. 
When sensor values cross predefined thresholds, the 
onboard system generates DTCs, structured fault 
indicators that name the affected component. Each 
modality encodes a different aspect of the same 
underlying fault, yet in current practice they are 
analyzed in complete 
isolation~\cite{theissler2021predictive,ran2019survey}. 
We study this problem using proprietary operational 
data from a major heavy-duty vehicle manufacturer's 
global service network.

Consider a concrete example: a technician writes 
``intermittent power loss under load.'' The DTC log 
shows P0299 (turbo underboost). The sensor readout 
shows anomalous boost pressure rankings. These 
three signals describe the same fault from different 
perspectives, yet no existing system connects them 
automatically. The question we investigate is 
whether fusing these modalities improves diagnostic 
classification or whether text, as the richest 
single source, already subsumes the information in 
sensors and DTCs.

Cross-modal alignment has succeeded in 
vision-language settings~\cite{radford2021learning} 
and time-series-language 
tasks~\cite{xue2024promptcast,li2025sensorllm}, 
but industrial vehicle diagnostics presents 
distinct challenges. Complaints span 10+ languages 
and are mostly non-diagnostic. Sensor data is over 
80\% missing due to hardware configuration 
differences across vehicle models. The temporal 
relationship between modalities is causal rather 
than correlational: a complaint is filed 
\textit{after} a failure, while sensors capture 
state \textit{before} it. Initial pilot experiments 
using retrieval and contrastive alignment confirmed 
these difficulties (Appendix~\ref{app:pilot}).

Our main experiment is a systematic ablation for 
engine component classification (5 classes, 885 
samples) evaluated with 5-fold cross-validation. We 
compare logistic regression, gradient-boosted 
trees, and MLP models across all modality 
combinations. Our contributions are:

\begin{enumerate}
\itemsep0em
\item The first application of three-way modality 
analysis (text, sensors, DTCs) to industrial 
vehicle diagnostics, with a modality dropout 
training strategy where text+DTC fusion achieves 
68.8\%, outperforming text-only (65.3\%) and all 
classical baselines.

\item Per-class evidence of complementarity: sensors dominate on intake/exhaust faults (93\%), fusion with modality dropout nearly triples fuel system accuracy (15\% to 38\%), and DTCs alone reach 58\% from a binary vector.
\end{enumerate}

\section{Related Work}

\subsection{Multi-Modal Contrastive Learning}

CLIP~\cite{radford2021learning} aligns images and 
text through contrastive pre-training on 400M 
pairs. Extensions to 
audio~\cite{guzhov2022audioclip} and inertial 
sensors~\cite{moon2023imu2clip} demonstrate the 
generality of the approach but rely on dense, 
regular signals and abundant paired data. 
Modality-level dropout, where entire input channels 
are randomly disabled during training, has been 
explored in audio-visual and medical multi-modal 
settings~\cite{neverova2015moddrop} to prevent 
modality dominance; we apply this strategy to 
industrial diagnostics for the first time.

\subsection{Sensor-Language Alignment}

PromptCast~\cite{xue2024promptcast} converts 
time-series forecasting into a sentence-to-sentence 
task. SensorLLM~\cite{li2025sensorllm} aligns 
motion sensor embeddings with text through a 
two-stage framework for human activity recognition 
on regular-frequency signals (50--100Hz). Our 
sensor data is structurally different: 
pre-aggregated ECU diagnostics read at irregular 
workshop visits, with over 80\% missing values. 
Neither approach is directly applicable.

\subsection{Industrial Text Mining and Maintenance}

Brundage et al.~\cite{brundage2021technical} survey 
NLP for manufacturing maintenance; Sexton et 
al.~\cite{sexton2017hybrid} propose hybrid methods 
for technician-generated text. These efforts 
process complaint text in isolation without 
connecting it to sensor data or fault codes. 
TEST~\cite{sun2024test} and 
Time-LLM~\cite{jin2024timellm} align temporal data 
with text representations but target forecasting 
with clean signals. Traditional predictive 
maintenance relies on supervised learning over 
labeled failure 
data~\cite{theissler2021predictive,ran2019survey,zhang2026pdmbench}. 
Retrieval-Augmented 
Generation~\cite{lewis2020retrieval} has been 
applied to technical manual 
retrieval~\cite{barnett2024seven}, but these 
frameworks remain sensor-blind, retrieving textual 
remedies without validating them against physical 
telemetry. Cross-modal diagnosis connecting textual 
symptoms to quantitative sensor evidence remains 
largely unaddressed.

\section{Data}
\label{sec:data}

Our dataset is from a major heavy-duty vehicle 
manufacturer and contains three modalities linked 
by vehicle identifier (VIN) and timestamp 
(Table~\ref{tab:data}).

\begin{table}[t]
\centering
\small
\caption{Dataset characteristics (approximate values 
due to proprietary constraints).}
\label{tab:data}
\begin{tabular}{@{}p{4.2cm}r@{}}
\toprule
\multicolumn{2}{@{}l}{\textbf{Complaint corpus}} \\
\midrule
Service records & $\sim$9K \\
Languages detected & 10+ \\
\quad EN / FR / DE / ES / PL & 31 / 18 / 8 / 7 / 6\% \\
Diagnostic complaints (est.) & $\sim$32\% \\
\midrule
\multicolumn{2}{@{}l}{\textbf{Sensor telemetry}} \\
\midrule
Readout rows & $>$700K \\
Sensor groups $\times$ features & 525 $\times$ 6 \\
Global NaN fraction & $>$80\% \\
\midrule
\multicolumn{2}{@{}l}{\textbf{Diagnostic Trouble Codes}} \\
\midrule
DTC records & $>$51M \\
Unique DTC codes & 1,677 \\
VINs with DTCs & 89\% of complaints \\
Mean DTCs per complaint (-30d) & 30 \\
\midrule
\multicolumn{2}{@{}l}{\textbf{Cross-modal overlap}} \\
\midrule
Vehicles with all 3 modalities & $\sim$74\% \\
Engine-domain triplets & 885 \\
\bottomrule
\end{tabular}
\end{table}

\textbf{Complaints} are service records written by 
technicians across global markets in 10+ languages. 
Approximately 68\% are non-diagnostic (parts 
requests, campaign notes), making the corpus noisy 
for fault classification.

\textbf{Sensors} consist of 525 diagnostic groups, 
each reporting six pre-aggregated features computed 
onboard by the ECU (fault counter, last ranking, 
ranking average, standard deviation, update 
counter, worst ranking). A missing value indicates 
that the corresponding diagnostic parameter was not 
updated during the readout period, resulting in 
structured sparsity where only 19.6\% of values are 
observed per sample.

\textbf{DTCs} are structured fault signals generated 
when sensor values cross predefined thresholds. The 
database contains over 51M records across 7,339 
vehicles with 1,677 unique codes. For each 
complaint, we collect all DTCs recorded for the 
same vehicle within last 30 day window and encode 
them as a multi-hot vector over the 500 most 
frequent codes.

\textbf{Triplet construction:} Aligning three 
independently maintained industrial databases into 
coherent triplets is a non-trivial data engineering 
challenge. Complaints, sensor readouts, and DTC 
logs are recorded by different systems, at 
different frequencies, with different coverage: 
complaints are filed at service visits, sensor 
readouts are captured during workshop diagnostic 
scans, and DTCs accumulate continuously in onboard 
memory. For each complaint, we retrieve the most 
recent sensor readout \textit{before} the complaint 
date for the same vehicle, ensuring the sensor 
snapshot reflects the state leading up to the 
fault. Only $\sim$74\% of vehicles in the complaint 
corpus have coverage in all three databases, and 
temporal alignment further reduces the yield. This 
produces 3,595 matched triplets from $\sim$9K 
original complaints, a 60\% attrition rate inherent 
to cross-database alignment in industrial settings. 
Of these, more than 99\% correspond to distinct vehicles 
(13 VINs contribute exactly two triplets each). The 
resulting dataset represents the full matchable 
population for this manufacturer, not a sample that 
could be enlarged by collecting more data.

\textbf{Engine component focus.} We restrict to 
engine-related complaints (functional group prefixes 
20--29) and exclude some classes due to insufficient 
samples. This yields 885 triplets across five 
classes: mechanical (21, $n$=187), fuel system (22, 
$n$=60), cooling (23, $n$=141), intake/exhaust (25, 
$n$=367), and electronics (28, $n$=130).

\section{Method}

\begin{figure*}[t]
\centering
\includegraphics[width=0.9\textwidth]{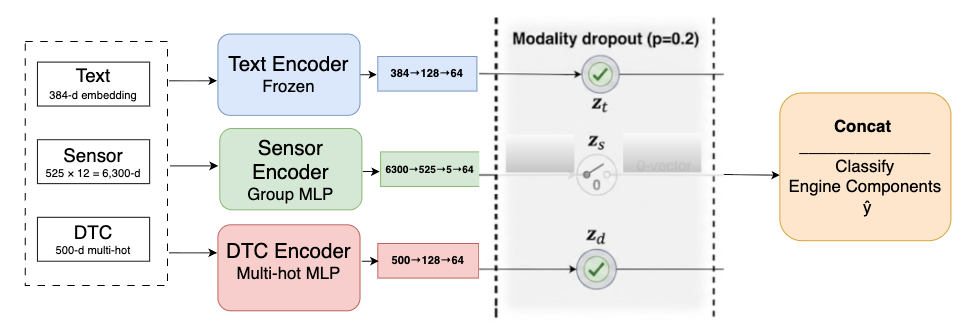}
\caption{Model architecture. Three parallel encoders 
produce 64-d embeddings from text (384-d sentence 
embedding), sensor (525 groups $\times$ 12 
features), and DTC (500-d multi-hot) inputs. During 
training, modality dropout independently zeros each 
branch with probability $p=0.2$ (illustrated here 
with the sensor branch disabled). At inference, all 
branches are active. The masked embeddings are 
concatenated and classified into five engine 
component classes.}
\label{fig:arch}
\end{figure*}

Given a complaint, a sensor readout, and a set of 
DTC codes associated with a vehicle visit, the task 
is to classify which of five engine component 
classes is affected. We describe the input 
representation for each modality, the fusion 
architecture, and the modality dropout strategy. 
Figure~\ref{fig:arch} shows the overall 
architecture.

\subsection{Input Representations}

\textbf{Text.} Each complaint is encoded by a 
frozen pre-trained sentence transformer 
(all-MiniLM-L6-v2), producing a 384-dimensional 
embedding $\mathbf{t} \in \mathbb{R}^{384}$. No 
fine-tuning is applied to the text encoder. We keep 
the encoder frozen to avoid overfitting on 885 
samples and to ensure the text representations 
remain general across the multilingual complaint 
corpus.

\textbf{Sensors.} For each complaint, we retrieve 
the most recent sensor readout before the complaint 
date. The raw input is a matrix 
$\mathbf{S} \in \mathbb{R}^{G \times 2F}$ where 
$G = 525$ sensor groups and $F = 6$ features. Each 
group contributes $F$ values concatenated with $F$ 
binary observation indicators (1 if the feature was 
updated during the readout period, 0 if it remained 
at its default NaN value), yielding 12 inputs per 
group. A per-group MLP $\phi_g$ compresses each 
group to a scalar health score:
\begin{equation}
h_i = \phi_g([\mathbf{s}_i; \mathbf{m}_i]) \in \mathbb{R}, \quad i = 1, \dots, G
\end{equation}
where $\mathbf{s}_i \in \mathbb{R}^F$ are the 
feature values and $\mathbf{m}_i \in \mathbb{R}^F$ 
is the observation mask. We then compute global 
statistics across observed groups:
\begin{equation}
\mathbf{v}_s = \left[\mu_h, \sigma_h, \max_h, \min_h, \frac{|\{i : \|\mathbf{m}_i\|_1 > 0\}|}{G}\right] \in \mathbb{R}^5
\end{equation}
where $\mu_h$ and $\sigma_h$ are the mean and 
standard deviation of health scores over groups 
with at least one observed feature. The final 
sensor representation is obtained via a two-layer 
MLP: $\mathbf{e}_s = \psi_s(\mathbf{v}_s) \in \mathbb{R}^{64}$.

\paragraph{DTCs:}
For each training fold, we select the 500 most frequent DTC codes using
training data only and apply the resulting vocabulary to the corresponding
validation fold. For each complaint, we collect all DTCs recorded for the
same vehicle within last 30-day window centered on the diagnostic service
visit, and encode them as a multi-hot vector
$\mathbf{d} \in \{0,1\}^{500}$. A two-layer MLP produces the DTC embedding:
$\mathbf{e}_d = \psi_d(\mathbf{d}) \in \mathbb{R}^{64}$.

\subsection{Fusion and Classification}

Each modality encoder produces a 64-dimensional 
embedding. The text embedding is obtained via 
$\mathbf{e}_t = \psi_t(\mathbf{t}) \in \mathbb{R}^{64}$. 
The concatenated representation feeds a classifier:
\begin{equation}
\hat{y} = f_\theta([\mathbf{e}_t; \mathbf{e}_s; \mathbf{e}_d]) \in \mathbb{R}^C
\end{equation}
where $C = 5$ is the number of engine component 
classes and $f_\theta$ is a three-layer MLP with 
ReLU activations and standard dropout ($p = 0.3$). 
The model is trained with cross-entropy loss.

For ablation, we disable modalities by omitting the 
corresponding embeddings from the concatenation. 
Single-modality models use a 64-d input; pairwise 
models use 128-d; the full model uses 192-d. The 
classifier architecture adjusts its input dimension 
accordingly. We deliberately use a simple MLP 
architecture rather than attention-based or gating 
fusion mechanisms, as our dataset is too small to 
reliably train more complex fusion strategies 
without overfitting.

\subsection{Modality Dropout}

Naive fusion (Section~\ref{sec:results}) does not 
consistently outperform text alone, because the 
model learns to rely on the strongest single 
modality and treats the others as noise. To address 
this, we apply modality 
dropout~\cite{neverova2015moddrop}: during 
training, each modality is independently zeroed out 
with probability $p_{\text{drop}} = 0.2$ per batch, 
with the constraint that at least one modality 
remains active. Formally, at each training step we 
sample binary masks 
$z_t, z_s, z_d \sim \text{Bernoulli}(1 - p_{\text{drop}})$ 
and compute:
\begin{equation}
\hat{y} = f_\theta([z_t \cdot \mathbf{e}_t;\; z_s \cdot \mathbf{e}_s;\; z_d \cdot \mathbf{e}_d])
\end{equation}
If $z_t = z_s = z_d = 0$, we force $z_t = 1$, as 
text has the highest standalone accuracy. At 
inference, all modalities are active 
($z_t = z_s = z_d = 1$). This forces the classifier 
to extract useful signal from every modality 
combination, preventing the network from ignoring 
weaker modalities. Because modality dropout trains 
the classifier to operate with any subset of 
modalities, the model naturally handles missing 
modalities at inference: if a vehicle lacks sensor 
data or DTC records, the corresponding branch is 
zeroed and the classifier produces a prediction 
from the available modalities without retraining.

\subsection{Training Details}

All MLP models are trained for 80 epochs with 
AdamW (learning rate $10^{-3}$, weight decay 
$10^{-4}$) and cosine annealing. Gradient norms are 
clipped to 1.0. Batch size is 32. Random seed is 
fixed at 42. We evaluate with stratified 5-fold 
cross-validation and report mean $\pm$ standard 
deviation.

For classical baselines, we use logistic regression 
(LR, $C = 1.0$, max 2000 iterations) and 
gradient-boosted trees (XGBoost, 200 trees, max 
depth 4, learning rate 0.1) on the same input 
features. Text features are the 384-d sentence 
embeddings. Since logistic regression and XGBoost 
operate on fixed-length feature vectors and cannot 
incorporate a learned per-group neural encoder, 
sensor features are instead a 13-d hand-crafted 
summary (mean and standard deviation of each of the 
6 raw feature types across observed groups, plus 
the overall observation fraction). DTC features are 
the 500-d multi-hot vector. For combined models, 
features are concatenated.

\section{Results}
\label{sec:results}

\begin{table}[t]
\centering
\small
\caption{Engine component classification (5 classes, 
885 samples, 5-fold CV). Best result in 
\textbf{bold}. $\times$ Rand.\ indicates the 
multiple over the 20.0\% uniform random baseline. 
Weighted F1 reported for key models.}
\label{tab:main}
\begin{tabular}{@{}llrrr@{}}
\toprule
Method & Modalities & Acc.\ (\%) & $\times$ Rand. & wF1 \\
\midrule
\multicolumn{5}{@{}l}{\textit{Baselines}} \\
Random & & 20.0 & 1.0 & -- \\
Majority & & 41.5 & 2.1 & -- \\
\midrule
\multicolumn{5}{@{}l}{\textit{Logistic Regression}} \\
LR & Sensor & 44.6 $\pm$ 1.5 & 2.2 & -- \\
LR & DTC & 54.8 $\pm$ 2.2 & 2.7 & -- \\
LR & Text & 62.8 $\pm$ 3.5 & 3.1 & -- \\
\midrule
\multicolumn{5}{@{}l}{\textit{Gradient-Boosted Trees}} \\
XGB & Sensor & 45.1 $\pm$ 2.4 & 2.3 & -- \\
XGB & DTC & 56.9 $\pm$ 1.5 & 2.8 & -- \\
XGB & Text & 61.9 $\pm$ 1.8 & 3.1 & -- \\
XGB & All & 65.5 $\pm$ 2.5 & 3.3 & -- \\
\midrule
\multicolumn{5}{@{}l}{\textit{MLP (ours)}} \\
MLP & Sensor & 47.0 $\pm$ 1.2 & 2.4 & -- \\
MLP & DTC & 58.0 $\pm$ 2.2 & 2.9 & -- \\
MLP & Sens.+DTC & 57.7 $\pm$ 0.7 & 2.9 & -- \\
MLP & Text & 65.3 $\pm$ 2.2 & 3.3 & .64 \\
MLP & Text+Sens. & 64.4 $\pm$ 3.3 & 3.2 & -- \\
MLP & Text+DTC & 67.9 $\pm$ 2.0 & 3.4 & .66 \\
MLP & All & 67.2 $\pm$ 1.8 & 3.4 & -- \\
MLP+MD & All & 67.9 $\pm$ 3.1 & 3.4 & -- \\
MLP+MD & Text+DTC & \textbf{68.8 $\pm$ 1.6} & 3.4 & \textbf{.67} \\
\bottomrule
\end{tabular}
\end{table}

\begin{figure}[t]
\centering
\includegraphics[width=0.8\columnwidth]{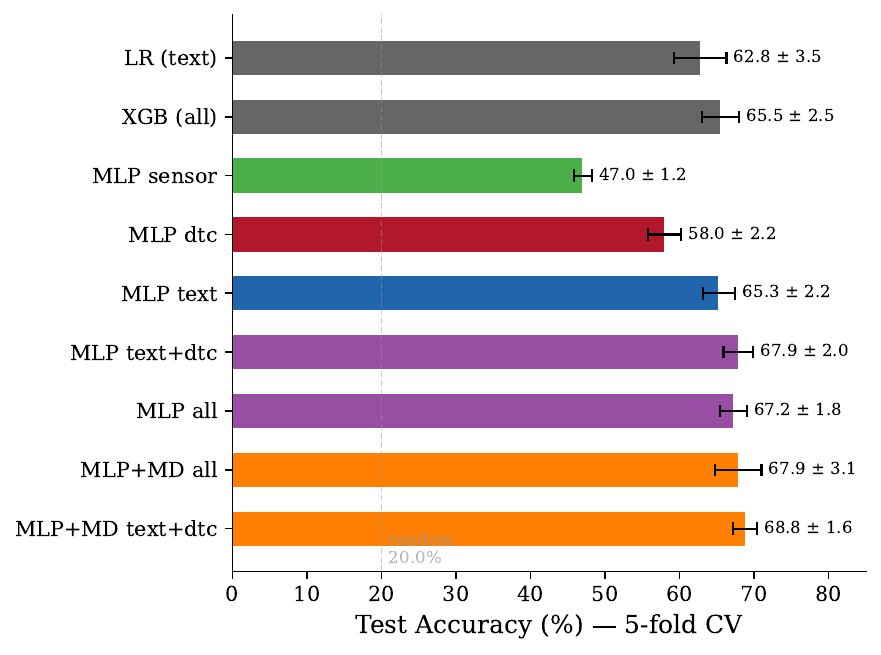}
\caption{Accuracy across methods and modality 
combinations (5-fold CV). Text+DTC with modality 
dropout achieves the highest accuracy, outperforming 
text-only MLP and all classical baselines.}
\label{fig:main}
\end{figure}

Table~\ref{tab:main} and Figure~\ref{fig:main} 
present the main results. The random baseline 
(20.0\%) reflects uniform prediction across five 
classes ($1/C = 1/5$). The weighted F1 score 
confirms the accuracy trend: text-only achieves 
0.64 while modality dropout fusion reaches 0.67, 
indicating improvement across both common and rare 
classes. Four findings emerge.

\subsection{Each Modality Carries Independent Signal}

Across all three model families, the ranking is 
consistent: text $>$ DTC $>$ sensor. Sensors alone 
reach 45--47\%, DTCs reach 55--58\%, and text 
reaches 62--65\%, all well above the 20.0\% random 
baseline. The consistency across LR, XGBoost, and 
MLP confirms this is a data property, not an 
artifact of a particular model.

\subsection{Naive Fusion Does Not Help}

MLP with all three modalities (67.2\%) modestly 
exceeds text-only (65.3\%). Adding sensors to text 
actually hurts (64.4\%) because the noisy, 
high-dimensional sensor embeddings introduce noise 
into the concatenated representation, diluting the 
text signal without contributing compensating 
information. This is consistent with findings in 
multi-modal learning where noisy modalities degrade 
the stronger signal~\cite{radford2021learning}.

\begin{figure}[t]
\centering
\includegraphics[width=0.8\columnwidth]{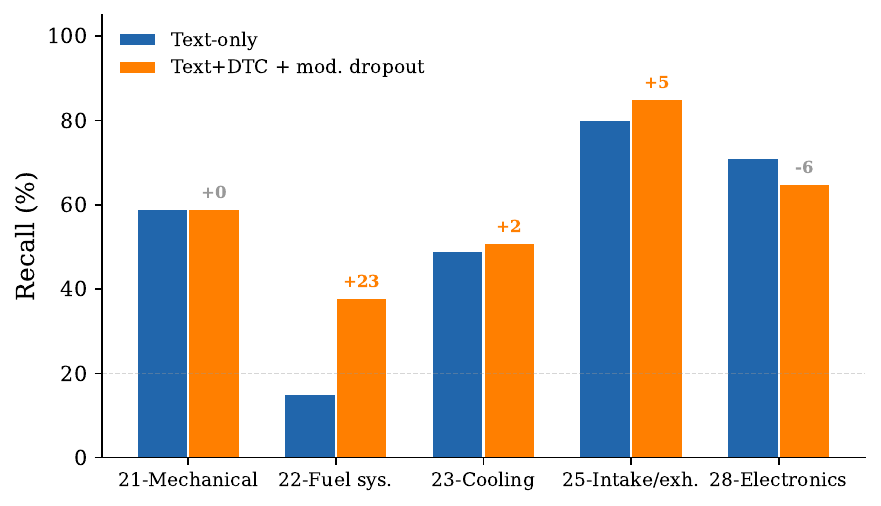}
\caption{Per-class recall comparison of text-only 
versus text+DTC with modality dropout. Fusion 
improves fuel system (+23pp), intake/exhaust 
(+5pp), and cooling (+2pp), with no change on 
mechanical and a decline on electronics ($-$6pp).}
\label{fig:delta}
\end{figure}

\subsection{Modality Dropout Makes Fusion Work}

\begin{figure}[t]
\centering
\includegraphics[width=0.8\columnwidth]{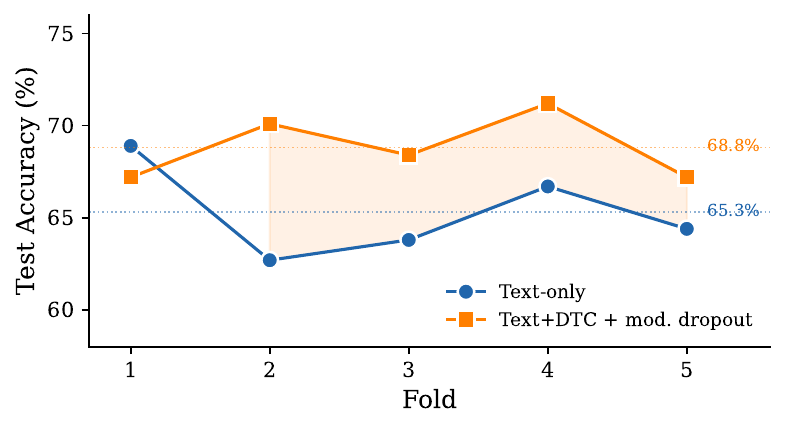}
\caption{Fold-by-fold accuracy: text+DTC with 
modality dropout (orange) versus text-only (blue) 
across 5-fold CV. Fusion outperforms text in 4 of 5 
folds with a mean improvement of 3.5 percentage 
points.}
\label{fig:folds}
\end{figure}

Text+DTC with modality dropout (68.8\%) achieves 
the best accuracy among all configurations, with 
low variance ($\pm$1.6\%). This outperforms 
text-only (65.3\%) by 3.5 points (paired $t$-test 
over 5 folds: $t=1.49$, $df=4$, $p=0.21$). While 
not statistically significant with 5 paired 
observations, modality dropout improves over 
text-only in 4 of 5 folds 
(Figure~\ref{fig:folds}). The three-modality model 
with dropout (67.9\%) does not improve over 
text+DTC+dropout, suggesting the sensor branch 
contributes limited additional signal. Modality 
dropout acts as an effective fusion regularizer 
that prevents dominant-modality collapse in both 
two-modality and three-modality settings.

\subsection{Text + DTC Is the Strongest Pair}

Among pairwise combinations, text + DTC (67.9\%) 
outperforms text + sensor (64.4\%) and sensor + DTC 
(57.7\%). DTCs add the most complementary signal to 
text, likely because DTCs name the affected 
component (structured) while text describes the 
symptom (unstructured). Sensors contribute least to 
pairwise fusion, possibly due to their 80\% 
missingness and the aggressive dimensionality 
reduction from 6,300 inputs to a 5-d summary, which 
may discard discriminative group-level patterns. A 
more expressive sensor encoder (e.g., attention 
over observed groups) could improve sensor 
contributions.

\subsection{Per-Class Analysis}

\begin{figure}[t]
\centering
\includegraphics[width=0.8\columnwidth]{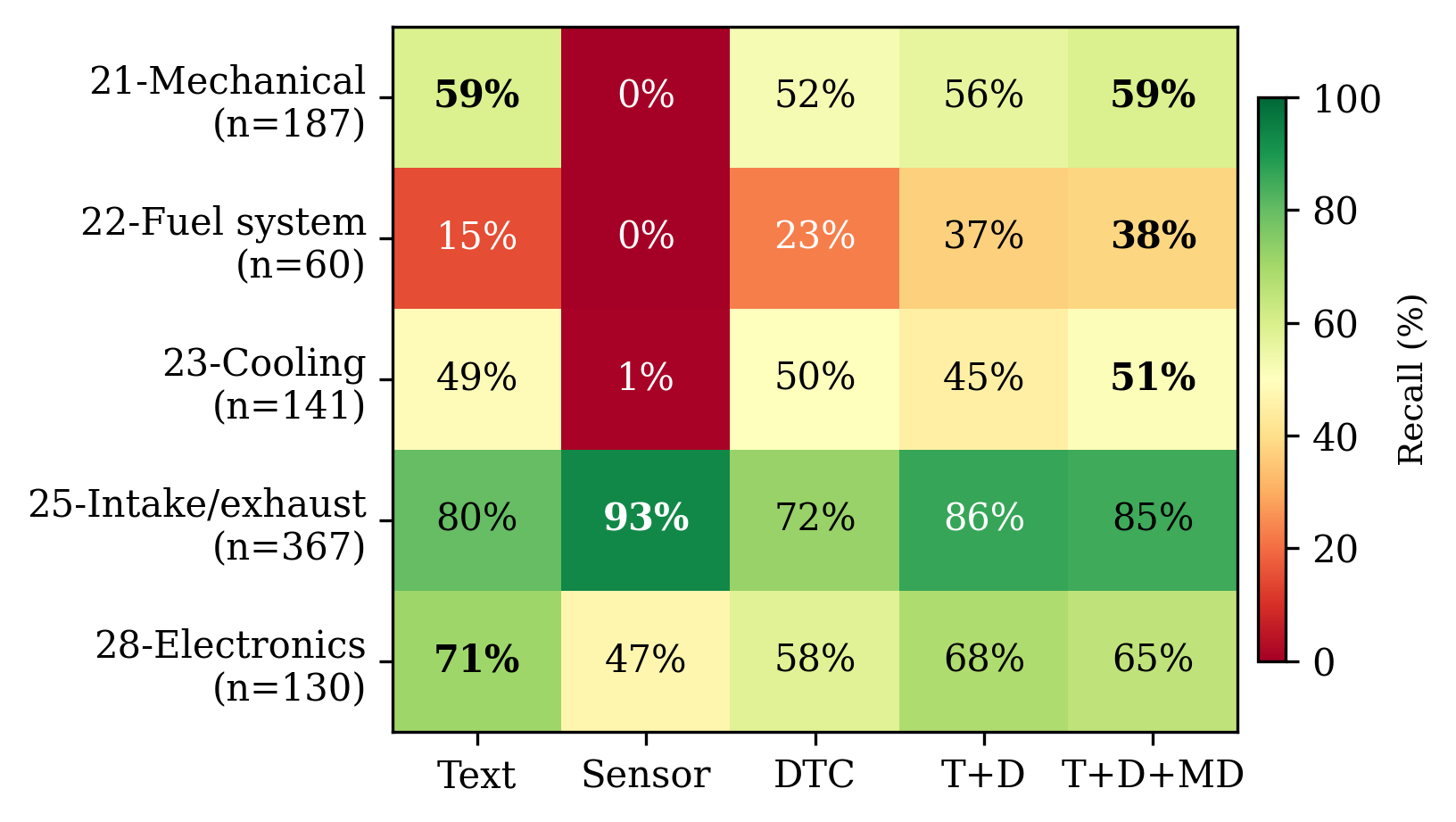}
\caption{Per-class recall by modality (pooled 
out-of-fold predictions across 5 folds). Bold 
values indicate the best result per class. The 
dominant modality varies by component.}
\label{fig:perclass}
\end{figure}

Figure~\ref{fig:perclass} reveals that the dominant 
modality varies by component class.

\textbf{Sensors dominate intake/exhaust faults.} On 
class 25 (intake/exhaust), sensors alone achieve 
93\%, outperforming text (80\%) and DTCs (72\%). 
Boost pressure rankings and air intake measurements 
directly quantify the physical condition that text 
can only describe qualitatively.

\textbf{Fusion rescues the fuel system class.} On 
class 22 (fuel system), text achieves only 15\% and 
sensors 0\%. The text+DTC model with modality dropout 
reaches 38\%, nearly tripling accuracy, because DTC codes 
like P0087 (fuel rail pressure too low) provide the 
structured signal that vague complaint text lacks.

\textbf{Cooling benefits from cross-modal synergy.} 
On class 23 (cooling), the T+D+MD model achieves 
51\% versus 49\% for text alone, a modest 
improvement. Neither modality is strong 
individually, but together with modality dropout 
they provide complementary evidence.

\textbf{Text suffices for electronics.} On class 28 
(electronics), text alone achieves 71\%, the 
highest single-modality result, because complaints 
like ``ECU fault code stored'' or ``wiring harness 
damage'' are already diagnostic. This class is 
inherently more diagnosable from text because 
electronic faults produce specific, unambiguous 
symptoms.

\section{Discussion}

Each modality captures a different diagnostic 
dimension. Text reflects how a technician 
\textit{perceives} a symptom. DTCs encode what the 
onboard system \textit{detected} based on sensor 
thresholds. Sensors measure the \textit{physical 
state}, including gradual degradation that may not 
yet trigger a DTC. The per-class results confirm 
this: sensors dominate where physical measurements 
are definitive (intake/exhaust), text dominates 
where the symptom description is most informative 
(electronics), and fusion helps where neither 
modality suffices alone (fuel system, cooling).

Without modality dropout, the classifier appears to rely predominantly on the strongest modality, limiting the benefit obtained from weaker branches. This is evidenced by naive fusion (67.2\%) only modestly outperforming text-only (65.3\%), whereas text+DTC+dropout achieves the best result (68.8\%). Modality dropout forces the network to learn from every modality subset, analogous to standard dropout~\cite{srivastava2014dropout} applied at the modality level. 
The fact that text+DTC+dropout outperforms all+dropout  (68.8\% vs.\ 67.9\%) indicates that additional modalities do not necessarily improve performance. More broadly, the per-class results suggest that the utility of each modality is fault-dependent.

The practical value of fusion lies in class-specific 
gains rather than aggregate improvement. Text-only 
achieves 15\% on fuel system faults (near random), 
while text+DTC reaches 37\%. DTCs alone achieve 
58.0\% from a raw binary vector that treats each 
code as an opaque identifier. However, DTC codes 
have internal structure: the first character 
indicates the system (P=powertrain, B=body, 
C=chassis, U=network) and subsequent digits encode 
the component and failure type. Encoding DTC 
textual descriptions (e.g., ``NOx Sensor Bank 1 
Sensor 2'') could capture semantic similarity 
between related codes and is a promising direction 
for future work.

\section{Conclusion}

We investigated three-way modality fusion for 
heavy-duty vehicle engine diagnostics. Each 
modality captures a different diagnostic dimension: 
text describes symptoms (65.3\%), DTCs encode 
structured faults (58.0\%), and sensors measure 
physical state (47.0\%). Naive fusion fails to 
improve over text, but modality dropout on text+DTC 
achieves 68.8\%, the best result across all 
methods, while the sensor branch provides limited 
additional value beyond text+DTC. Per-class 
analysis reveals genuine complementarity: sensors 
dominate intake/exhaust faults (93\%), fusion 
nearly triples fuel system accuracy (15\% to 38\%), 
and the combined model consistently improves on 
classes where text alone is weakest. These findings 
demonstrate that multi-modal fusion for industrial 
diagnostics requires training strategies that 
prevent modality collapse, and that the practical 
value of fusion lies in class-specific gains rather 
than aggregate accuracy improvement.

\section*{Limitations}

Our dataset contains 885 engine-domain triplets 
across five classes and exhibits class imbalance 
(ranging from 60 to 367 samples). While small for 
machine learning, this represents the full matchable 
population for this manufacturer, constrained by 
the requirement to temporally align three 
independent databases where only 74\% of vehicles 
have complete coverage. Results reflect one truck 
manufacturer's data and may not generalize to other 
OEMs or vehicle types. The text encoder 
(all-MiniLM-L6-v2) is English-centric; a 
multilingual encoder may yield different modality 
rankings given the 10+ languages in the corpus. The 
DTC window (last 30 days) and modality dropout 
probability (0.2) were chosen pragmatically rather 
than optimized, and alternative dropout strategies 
were not explored. Furthermore, we used a simple 
concatenation fusion method; investigating dynamic 
gating mechanisms to weight modalities per sample 
remains future work.

\bibliographystyle{unsrt}
\bibliography{custom}

@article{lewis2020retrieval,
  title={Retrieval-augmented generation for knowledge-intensive nlp tasks},
  author={Lewis, Patrick and Perez, Ethan and Piktus, Aleksandra and Petroni, Fabio and Karpukhin, Vladimir and Goyal, Naman and K{\"u}ttler, Heinrich and Lewis, Mike and Yih, Wen-tau and Rockt{\"a}schel, Tim and others},
  journal={Advances in neural information processing systems},
  volume={33},
  pages={9459--9474},
  year={2020}
}

@inproceedings{radford2021learning,
  title={Learning transferable visual models from natural language supervision},
  author={Radford, Alec and Kim, Jong Wook and Hallacy, Chris and Ramesh, Aditya and Goh, Gabriel and Agarwal, Sandhini and Sastry, Girish and Askell, Amanda and Mishkin, Pamela and Clark, Jack and others},
  booktitle={International conference on machine learning},
  pages={8748--8763},
  year={2021},
  organization={PmLR}
}

@inproceedings{barnett2024seven,
  title={Seven failure points when engineering a retrieval augmented generation system},
  author={Barnett, Scott and Kurniawan, Stefanus and Thudumu, Srikanth and Brannelly, Zach and Abdelrazek, Mohamed},
  booktitle={Proceedings of the IEEE/ACM 3rd International Conference on AI Engineering-Software Engineering for AI},
  pages={194--199},
  year={2024}
}

@article{theissler2021predictive,
  title={Predictive maintenance enabled by machine learning: Use cases and challenges in the automotive industry},
  author={Theissler, Andreas and P{\'e}rez-Vel{\'a}zquez, Judith and Kettelgerdes, Marcel and Elger, Gordon},
  journal={Reliability engineering \& system safety},
  volume={215},
  pages={107864},
  year={2021},
  publisher={Elsevier}
}

@article{ran2019survey,
  title={A survey of predictive maintenance: Systems, purposes and approaches},
  author={Zhu, Tianwen and Ran, Yongyi and Zhou, Xin and Wen, Yonggang},
  journal={arXiv preprint arXiv:1912.07383},
  year={2019}
}

@article{zhang2026pdmbench,
  title={PDMBench: A Standardized Platform for Predictive Maintenance Research},
  author={Zhang, Shuaicheng and Wang, Tuo and Kulkarni, Adithya and Adams, Stephen and Bhattacharya, Sanmitra and Tiyyagura, Sunil Reddy and Bowen, Edward and Veeramani, Balaji and Zhou, Dawei},
  year={2025}
}

@inproceedings{guzhov2022audioclip,
  title={Audioclip: Extending clip to image, text and audio},
  author={Guzhov, Andrey and Raue, Federico and Hees, J{\"o}rn and Dengel, Andreas},
  booktitle={ICASSP 2022-2022 IEEE International Conference on Acoustics, Speech and Signal Processing (ICASSP)},
  pages={976--980},
  year={2022},
  organization={IEEE}
}

@article{xue2024promptcast,
  title={Promptcast: A new prompt-based learning paradigm for time series forecasting},
  author={Xue, Hao and Salim, Flora D},
  journal={IEEE Transactions on Knowledge and Data Engineering},
  volume={36},
  number={11},
  pages={6851--6864},
  year={2023},
  publisher={IEEE}
}

@inproceedings{li2025sensorllm,
  title={Sensorllm: Aligning large language models with motion sensors for human activity recognition},
  author={Li, Zechen and Deldari, Shohreh and Chen, Linyao and Xue, Hao and Salim, Flora D},
  booktitle={Proceedings of the 2025 Conference on Empirical Methods in Natural Language Processing},
  pages={354--379},
  year={2025}
}

@inproceedings{moon2023imu2clip,
  title={IMU2CLIP: language-grounded motion sensor translation with multimodal contrastive learning},
  author={Moon, Seungwhan and Madotto, Andrea and Lin, Zhaojiang and Saraf, Aparajita and Bearman, Amy and Damavandi, Babak},
  booktitle={Findings of the Association for Computational Linguistics: EMNLP 2023},
  pages={13246--13253},
  year={2023}
}

@article{brundage2021technical,
  title={Technical language processing: Unlocking maintenance knowledge},
  author={Brundage, Michael P and Sexton, Thurston and Hodkiewicz, Melinda and Dima, Alden and Lukens, Sarah},
  journal={Manufacturing Letters},
  volume={27},
  pages={42--46},
  year={2021},
  publisher={Elsevier}
}

@inproceedings{sexton2017hybrid,
  title={Hybrid datafication of maintenance logs from AI-assisted human tags},
  author={Sexton, Thurston and Brundage, Michael P and Hoffman, Michael and Morris, Katherine C},
  booktitle={2017 ieee international conference on big data (big data)},
  pages={1769--1777},
  year={2017},
  organization={IEEE}
}

@inproceedings{sun2024test,
  title={Test: Text prototype aligned embedding to activate llm's ability for time series},
  author={Sun, Chenxi and Li, Hongyan and Li, Yaliang and Hong, Shenda},
  booktitle={International Conference on Learning Representations},
  volume={2024},
  pages={37854--37881},
  year={2024}
}

@inproceedings{jin2024timellm,
  title={Time-llm: Time series forecasting by reprogramming large language models},
  author={Jin, Ming and Wang, Shiyu and Ma, Lintao and Chu, Zhixuan and Zhang, James and Shi, Xiaoming and Chen, Pin-Yu and Liang, Yuxuan and Li, Yuan-Fang and Pan, Shirui and others},
  booktitle={International conference on learning representations},
  volume={2024},
  pages={23857--23880},
  year={2024}
}

@article{srivastava2014dropout,
  title={Dropout: a simple way to prevent neural networks from overfitting},
  author={Srivastava, Nitish and Hinton, Geoffrey and Krizhevsky, Alex and Sutskever, Ilya and Salakhutdinov, Ruslan},
  journal={The journal of machine learning research},
  volume={15},
  number={1},
  pages={1929--1958},
  year={2014},
  publisher={JMLR. org}
}

@article{neverova2015moddrop,
  title={Moddrop: adaptive multi-modal gesture recognition},
  author={Neverova, Natalia and Wolf, Christian and Taylor, Graham and Nebout, Florian},
  journal={IEEE Transactions on Pattern Analysis and Machine Intelligence},
  volume={38},
  number={8},
  pages={1692--1706},
  year={2015},
  publisher={IEEE}
}

\appendix

\section{Pilot Studies}
\label{app:pilot}

Before the main experiment, two pilot studies 
informed our approach.

\subsection{Decoupled Retrieval}

A RAG pipeline retrieves the top-5 most similar 
historical complaints using a pre-trained sentence 
transformer (all-MiniLM-L6-v2), while independently 
computing z-score anomalies for each sensor group. 
A domain expert evaluated five representative cases 
(Table~\ref{tab:expert}), rating each retrieved 
complaint on a 4-point relevance scale and 
classifying flagged sensors as Directly Diagnostic, 
Consequential (downstream effect), or Not Relevant. 
In 4 of 5 cases, at least one retrieved complaint 
was rated as partially or highly relevant. However, 
no flagged sensor was rated \textit{directly 
diagnostic}. Flagged sensors showed downstream 
effects rather than root causes because the z-score 
operates across the full signal space without 
knowing which sensors are relevant to the specific 
complaint. Case~3 yielded no relevant retrievals 
and was identified as a previously unseen failure 
type.

\begin{table}[h]
\centering
\small
\caption{Expert evaluation of decoupled retrieval. 
Relevance: 3=High, 2=Partial, 1=Not Relevant, 
0=Cannot Assess.}
\label{tab:expert}
\begin{tabular}{@{}lcccccc@{}}
\toprule
Case & Ret-1 & Ret-2 & Ret-3 & Ret-4 & Ret-5 & Sensors \\
\midrule
1 & 3 & 3 & 3 & 2 & 1 & Not Rel. \\
2 & 3 & 3 & 3 & 2 & 1 & Conseq. \\
3 & 1 & 1 & 1 & 1 & 1 & Not Rel. \\
4 & 3 & 3 & 2 & 1 & 1 & Conseq. \\
5 & 3 & 1 & 1 & 1 & 0 & Conseq. \\
\bottomrule
\end{tabular}
\end{table}

\subsection{Contrastive Alignment}

Following CLIP~\cite{radford2021learning}, we 
trained a dual-encoder (frozen text encoder + custom 
sensor GRU with 861K trainable parameters) with 
symmetric InfoNCE loss to embed complaints and 
sensor data into a shared 128-d space. 
Table~\ref{tab:contrastive} shows the cross-modal 
retrieval results on 553 test pairs. The model 
achieved 13$\times$ improvement over random at 
Top-1, with mean rank 164/553 (top 30\%). However, 
diagnostic complaints (31.5\%) achieved mean rank 
171 while non-diagnostic complaints (68.5\%) 
achieved 160, indicating the model learned 
vehicle-level correlations rather than fault-level 
alignment.

\begin{table}[h]
\centering
\small
\caption{Contrastive cross-modal retrieval on 553 
test pairs.}
\label{tab:contrastive}
\begin{tabular}{@{}lrrr@{}}
\toprule
Top-$k$ & Accuracy & Random & Improv. \\
\midrule
1 & 2.35\% & 0.18\% & 13.0$\times$ \\
5 & 6.15\% & 0.90\% & 6.8$\times$ \\
10 & 10.49\% & 1.81\% & 5.8$\times$ \\
50 & 30.92\% & 9.04\% & 3.4$\times$ \\
\bottomrule
\end{tabular}
\end{table}

These studies motivated two decisions: (1) move 
from instance-level retrieval to category-level 
classification, which pools weak per-instance signal 
into stronger per-category patterns; and (2) 
incorporate DTCs as a third modality providing 
structured supervision between text and sensors.

\end{document}